\pdfoutput=1
\documentclass[preprint,12pt,authoryear]{elsarticle}

\usepackage{amssymb}
\usepackage{adjustbox}
\usepackage{float}
\usepackage{color,soul}
\usepackage{aliascnt}
\usepackage{amsmath}
\usepackage{textgreek}
\usepackage{caption}
\usepackage{tikz-cd}
\usepackage{graphicx}
\usepackage{multirow}
\restylefloat{table}

\DeclareCaptionType{mycapequ}[][List of equations]
\newaliascnt{eqfloat}{equation}
\newfloat{eqfloat}{h}{eqflts}
\floatname{eqfloat}{Equation}
\newcommand*{\ORGeqfloat}{}
\let\ORGeqfloat\eqfloat
\def\eqfloat{%
  \let\ORIGINALcaption\caption
  \def\caption{%
    \addtocounter{equation}{-1}%
    \ORIGINALcaption
  }%
  \ORGeqfloat
}

\journal{Robotics and Computer-Integrated Manufacturing}

\begin{document}

\begin{frontmatter}



\title{Reframing demand forecasting: a two-fold approach for lumpy and intermittent demand}

\author[1,2,3]{Jo\v{z}e M. Ro\v{z}anec\corref{cor1}}
\ead{joze.rozanec@ijs.si}
\cortext[cor1]{Corresponding author.}

\author[1]{Dunja Mladeni\'{c}}
\ead{dunja.mladenic@ijs.si}

\address[1]{Jo\v{z}ef Stefan Institute, Jamova 39, 1000 Ljubljana, Slovenia}
\address[2]{Qlector d.o.o., Rov\v{s}nikova 7, 1000 Ljubljana, Slovenia}
\address[3]{Jo\v{z}ef Stefan International Postgraduate School, Jamova 39, 1000 Ljubljana, Slovenia}

\begin{abstract}
Demand forecasting is a crucial component of demand management. While shortening the forecasting horizon allows for more recent data and less uncertainty, this frequently means lower data aggregation levels and a more significant data sparsity. Sparse demand data usually results in lumpy or intermittent demand patterns, which have sparse and irregular demand intervals. Usual statistical and machine learning models fail to provide good forecasts in such scenarios. Our research shows that competitive demand forecasts can be obtained through two models: predicting the demand occurrence and estimating the demand size. We analyze the usage of local and global machine learning models for both cases and compare results against baseline methods. Finally, we propose a novel evaluation criterion of lumpy and intermittent demand forecasting models' performance.
Our research shows that global classification models are the best choice when predicting demand event occurrence. When predicting demand sizes, we achieved the best results using Simple Exponential Smoothing forecast. We tested our approach on real-world data consisting of 516 three-year-long time series corresponding to European automotive original equipment manufacturers' daily demand.
\end{abstract}

\begin{graphicalabstract}
\includegraphics[width=\textwidth,keepaspectratio]{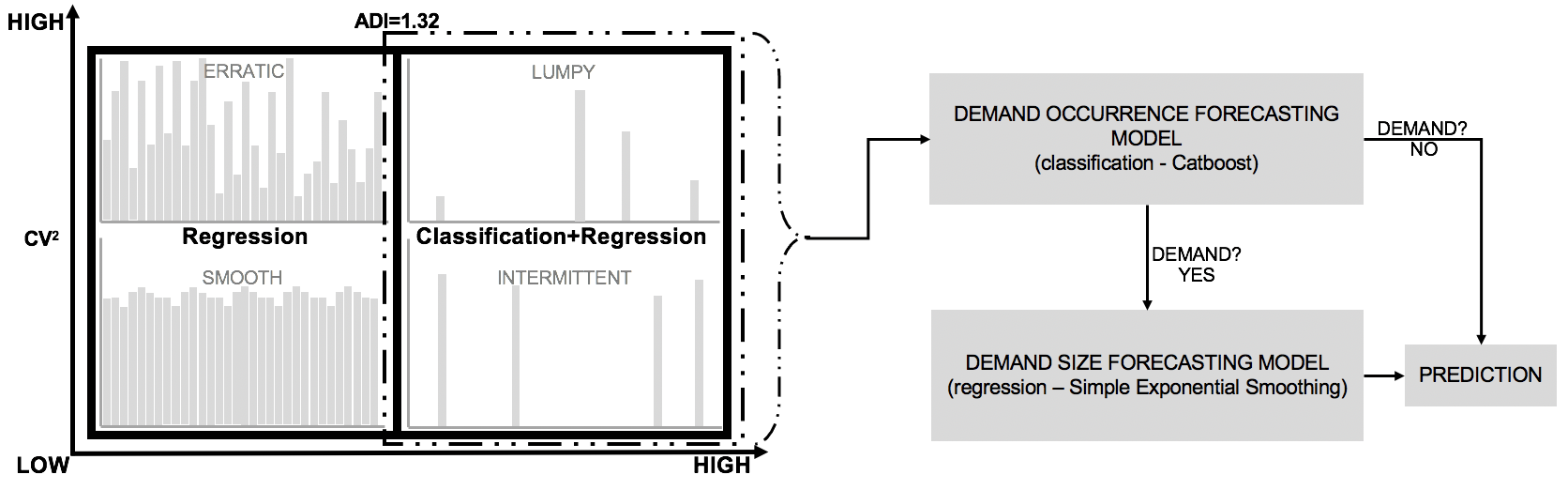}
\end{graphicalabstract}

\begin{highlights}
\item  Formulating demand occurrence and size estimation as separate forecasting problems
\item Evaluating performance with classification, regression, and inventory metrics
\item New model achieving state of the art results and testing it on a real-world dataset
\item Proposing a new demand classification schema
\end{highlights}


\end{frontmatter}


\section{Introduction}\label{section-intro}
Demand forecasting is a critical component of supply chain management, directly affecting production planning and order fulfillment. Accurate forecasts have an impact across the whole supply chain and manufacturing plant organization: operational and strategic decisions are made on resources (allocation and scheduling of raw material and tooling), workers (scheduling, training, promotions, or hiring), manufactured products (market share increase, production diversification) and logistics for deliveries.

To issue accurate forecasts, we have to consider demand characteristics. Multiple demand characterizations were proposed (\cite{williams1984stock,johnston1996forecasting}). Among the most influential ones is the characterization proposed by \cite{syntetos2005categorization}, which divides demand patterns into four quadrants, based on inter-demand interval and coefficient of variation. The four demand types are smooth (regular demand occurrence and low demand quantity variation), erratic (regular demand occurrence and high demand quantity variation), intermittent (irregular demand occurrence and low demand quantity variation), and lumpy (irregular demand occurrence and high demand quantity variation).
Intermittent and lumpy demand forecasting is considered among the most challenging demand forecasting problems. Both present infrequent demand arrivals with many zero demand periods, which pose an additional challenge to the accurate demand quantity estimation. Demand quantity estimation is harder for lumpy demands, since it also presents variable demand sizes (\cite{petropoulos2015forecast,amin2008neural,Bartezzaghi2011}).

\cite{bartezzaghi1999simulation} considers demand lumpiness as a consequence of different market characteristics, such as numerousness and heterogeneity of customers, the frequency at which customers place the orders, and the variety of customer's requests (e.g., high customization in make-to-order settings (\cite{verganti1997order})) and the correlation between customers behavior). Lumpyness is also related to the granularity level at which demand is considered (e.g., visualize demand at a client and product level vs. only at a product level) or visualize daily demand vs. monthly). Higher aggregation levels usually reduce the number of periods without demand, changing the demand pattern classification.

\cite{babai2014intermittent} stated that intermittent demand items account for considerable proportions of any organization's stock value. The importance of lumpy demand was characterized in a use case by \cite{johnston2003examination}, who found that 75\% of items had a lumpy demand, accounting for 40\% of the company's revenue and 60\% of stock investment. In this line, \cite{amin2008neural} cites multiple authors who observe lumpy patterns are widespread, especially in organizations that hold many spare parts, such as process and automotive industries, telecommunication systems, and others.

Increasing industry automation, digitalization, and information sharing (e.g., thorough Electronic Data Interchange software), fomented by national and regional initiatives  (\cite{davies2015briefing,glaser2019made,yang2021industry}), accelerates the data and information flow within the organization, enabling greater agility. It is critical to develop demand forecasting models capable of providing forecasts at a low granularity level to achieve greater agility in supply chain management. Such models enable short forecasting horizons and provide insights at a significant detail level, allowing to foresee and react to changes quickly. While these forecasting models benefit from the most recent data available (which helps enhance the forecast's accuracy), the low granularity level frequently requires dealing with irregular demand patterns.

Given the variety of demand types, researchers proposed multiple approaches to provide accurate demand forecasts. While smooth and erratic demands achieve good results using regression models, intermittent and lumpy demand require specialized models that consider demand occurrence. Statistical, machine learning and hybrid models were developed to that end. The increasing digitalization of industry enables to timely collect data relevant to demand forecasts. Data availability is key for developing machine learning models, which in some cases achieve the best results.

To deal with intermittent demand, \cite{croston1972forecasting} proposed a forecasting model that provides separate estimates for demand occurrence and demand quantity estimation. Since then, much work followed this direction: multiple authors proposed corrections to Croston's method to address forecast biases or provide different means to estimate demand occurrence.

The measurement of intermittent and lumpy model's performance was also the subject of extensive research. Many authors agree we require regression accuracy and inventory metrics. There is increasing agreement that regression metrics alone, used for smooth and erratic demand, are not useful for measuring intermittent and lumpy demand since they fail to weigh zero demand periods. Inventory metrics suffer the same bias while providing a perspective of how much time products stay in stock.

Croston provided a valuable intuition on separating demand occurrence from demand sizes. While many authors followed this intuition, we found that in the literature we reviewed, no author fully considered demand forecasting as a compound problem, which requires not only separate models but also separate metrics. We propose reframing demand forecasting as a two-phase problem, which requires (i) a classification model to predict demand occurrence and (ii) a regression model to predict demand sizes. For smooth and erratic demands, classification can be omitted since it (almost) always occurs. In those cases, using only a regression model provides good demand forecasts (\cite{bruhl2009sales,wang2011using,sharma2012sales,gao2018chinese,salinas2020deepar,bandara2020forecasting}). For intermittent and erratic demands, using separate models for classification and regression provides at least two benefits. First, separate models allow optimizing for different objectives. Second, each problem has adequate metrics, and the cause of performance / under-performance can be clearly understood and addressed.

The classification model, used to estimate demand occurrence, should use specific features that correlate to demand occurrence. We found past research is a rich source of intuitions on factors related to demand occurrence. We consider the classification model a global model that trained over all the time series or subsets. This way, even though demand events for a single product may be scarce, the greater the amount for products considered, demand events sparsity decreases. Simultaneously, the model can learn underlying patterns, which may be related to specific behaviors (e.g., deliveries take place only on certain days). When demand events are scarce, we may find the classification problem corresponds to an imbalanced one, posing an additional challenge.

In this research we propose:
\begin{enumerate}
 \item decouple the demand forecasting problem into two separate problems: classification (demand occurrence), and regression (demand quantity estimation);
 \item use four measurements to assess demand forecast performance: (i) area under the receiver operating characteristic curve (AUC ROC, see \cite{BRADLEY19971145}) to assess demand occurrence, (ii) two variations of the Mean Scaled Error (MASE, see \cite{hyndman2006another}) to assess demand quantity forecasts, and (iii) Stock-keeping-oriented Prediction Error Costs (SPEC), proposed by \cite{martin2020new}, as inventory metric;
 \item a new demand classification schema based on previous literature and our research findings
\end{enumerate}

We compare the statistical methods proposed by \cite{croston1972forecasting}, \cite{syntetos2005accuracy}), and \cite{teunter2009bias}, the hybrid models developed by \cite{nasiri2008hybrid} and \cite{willemain2004new}, and the ADIDA forecasting method introduced by \cite{nikolopoulos2011aggregate}, measuring their performance through classification, regression and inventory metrics. We also developed a compound model of our own, which outperforms the listed ones in all three dimensions.

We perform our research on a dataset consisting of 516 time series of intermittent and lumpy demand at a daily aggregation level, which corresponds to the demand of European manufacturing companies related to the automotive industry.

The rest of this paper is structured as follows: Section~\ref{RELATED-WORK} presents related work, Section~\ref{DEMAND-FORECASTING-REFRAMED} describes our approach to demand forecasting with a particular focus on intermittent and lumpy demands,  Section~\ref{METHODOLOGY} describes the features we created for each forecasting model, how we built and evaluated them, Section~\ref{EXPERIMENTS-AND-RESULTS} describes the experiments we performed and results obtained, and in Section~\ref{CONCLUSIONS}, we provide our conclusions and outline future work.

\section{Related work}\label{RELATED-WORK}
\subsection{Demand characterization}\label{RW-DEMAND-CHARACTERIZATION}

Many authors tried to characterize demand to provide cues to decide which forecasting model is most appropriate for each case. \cite{croston1972forecasting} assessed demand based on demand size and inter-demand intervals, providing a method to forecast intermittent demand. \cite{williams1984stock} considered variance of the number of orders and their size given a particular lead time, classifying items into five categories regarding high/low demand sporadicity and size. A particular category is created for products with a highly sporadic demand occurrence and high demand size variance, based on the author's empirical findings regarding demand intermittence. \cite{eaves2004forecasting} found the classification schema proposed by Williams did not provide means to distinguish steady demand solely based on transaction variability. He proposed dividing demand into five categories considering lead time variability, transaction rate variability, and demand size variability.

\begin{eqfloat}
\begin{equation}\label{EQ:ADI}
    ADI = \frac{\textit{Total Periods}}{\textit{Total Demand Buckets}}
\end{equation}
\caption{ADI stands for Average Demand Interval.}
\end{eqfloat}

\begin{eqfloat}
\begin{equation}\label{EQ:CV}
    CV = \frac{\textit{Demand Standard Deviation}}{\textit{Demand Mean}}
\end{equation}
\caption{CV stands for Coefficient of Variation. Is computed over non-zero demand occurrences (\cite{syntetos2001bias}).}
\end{eqfloat}

\begin{figure*}[!ht]
\centering
\includegraphics[width=5.0in]{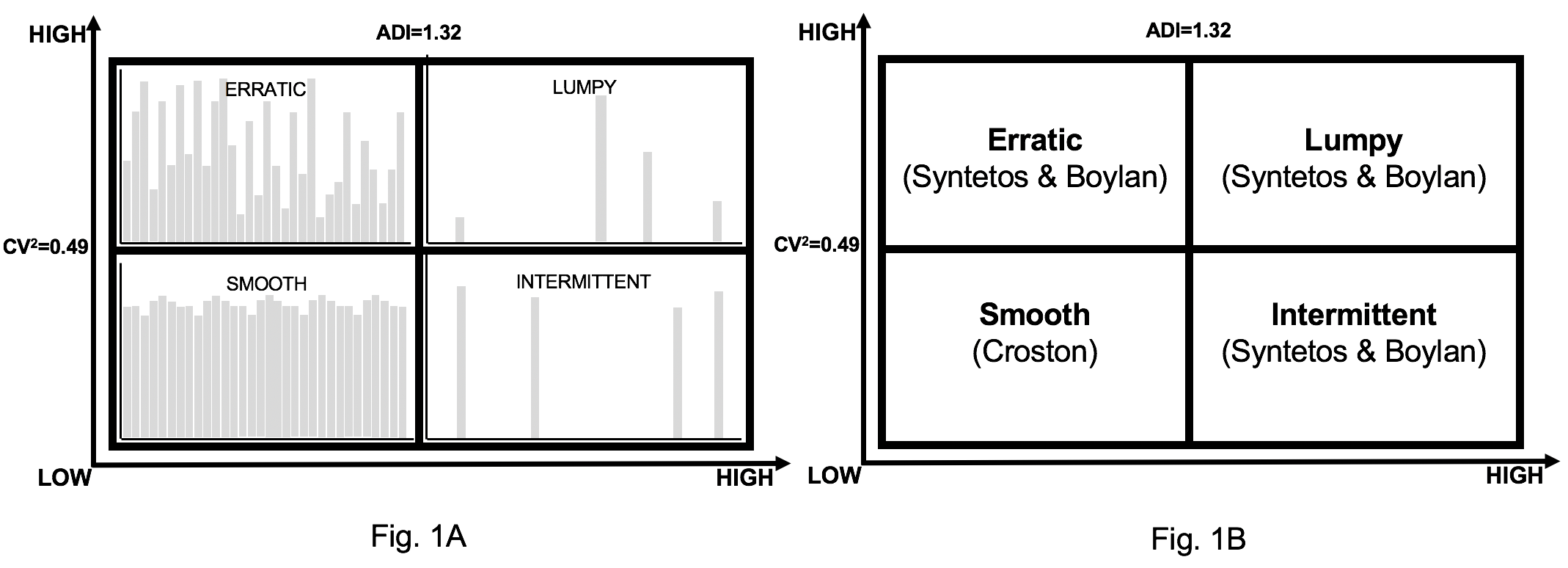}
\caption{Demand patterns classification. Fig. 1A shows classification proposed by \cite{syntetos2005categorization} based on empirical findings. In Fig. 1B we propose a new demand pattern classification, based on models required to solve the demand forecasting problem. 'R' stands for 'Regression', while 'C+R' stands for 'Classification + Regression'.}
\label{F:DEMAND-CLASSIFICATION}
\end{figure*}

\cite{johnston1996forecasting} introduced the concept of average demand interval (ADI, see Eq.~\ref{EQ:ADI}), which was complemented by \cite{syntetos2005categorization} introducing the coefficient of variation (CV, see Eq.~\ref{EQ:CV}). Both concepts allow to divide demand into quadrants, considering them smooth, erratic, intermittent, or lumpy demand types (see Fig.~\ref{F:DEMAND-CLASSIFICATION}). Smooth and erratic demand present regular demand, with smooth demand having little variability in demand sizes, while this variability is strong for erratic demands. Intermittent and lumpy demand present irregular demand intervals over time. Intermittent demand has little variability in demand sizes, in contrast to lumpy demand, which has a greater demand size variability. Thresholds were set based on empirical findings regarding where the methods proposed by \cite{croston1972forecasting} and \cite{syntetos1999correcting,syntetos2001bias} performed best.

This paper we use the term steady demand to refer to smooth and erratic demands (that display regular demand occurrence) and irregular demand for intermittent and lumpy demand (that display irregular demand occurrence).

\subsection{Demand Forecasting models}\label{RW-DEMAND-FORECASTING-MODELS}

Forecasting irregular demand is considered a challenging task since, in addition to demand size forecast, it requires taking into account irregular demand occurrence. We categorize demand forecasting models into three types. Type I consist of a single model providing a demand size estimate. Type II uses aggregation to remove demand intermittency and benefit from regular time-series models to forecast demand. Type III use separate models to estimate demand occurrence and demand size.

Box–Jenkins approaches, frequently used for regular time series forecasting, are considered useless in the context of irregular demand (\cite{wallstrom2010evaluation}) since it is challenging to estimate trends and seasonality given the high proportion of zeros.

\subsubsection{Type I models}\label{RW-TYPE-I}

\cite{wright1986forecasting} developed the linear, exponential smoothing, an adaptation of Holt's double exponential smoothing model, which considers variable reporting frequency and irregularities in time spacing, to compute and update a trend line with exponential smoothing. \cite{altay2008adapting} demonstrated this method is useful to forecast intermittent demand, where the trend is present. \cite{sani1997selecting,ghobbar2003evaluation} found averaging methods can provide acceptable performance in some cases, despite demand intermittency. \cite{chatfield2007all} suggested using a zero demand for highly lumpy demands, where the holding cost is much higher than the shortage cost. \cite{gutierrez2008lumpy} proposed forecasting lumpy demand with a three-layer multilayer perceptron (MLP), considering only two inputs: demand of the immediately preceding period and the number of periods separating the last two non-zero demand transactions.

\begin{eqfloat}
\begin{equation}\label{EQ:CROSTON}
if d_t>0
\begin{cases}
    a_{t+1} = \alpha \cdot d_t + (1-\alpha)\cdot a_t\\
    p_{t+1} = \alpha \cdot q \cdot d_t + (1-\alpha)\cdot p_t\\
    f_{t+1} = \frac{a_t}{p_t}
\end{cases}
if d_t=0
\begin{cases}
    a_{t+1} = a_t\\
    p_{t+1} = p_t\\
    f_{t+1} = f_t
\end{cases}
\end{equation}
\caption{Croston's formula (\cite{croston1972forecasting}) for irregular demand estimation, where \textit{a} is demand level, \textit{p} is periodicity, \textit{d} refers to demand observations, \textit{q} is previous demand occurrence, and \textalpha represents a smoothing constant.}
\end{eqfloat}

\begin{eqfloat}
\centering
\begin{equation}\label{EQ:SBA}
Y_t = (1-\frac{\alpha}{2}) \cdot pred_{Croston}
\end{equation}
\caption{\cite{syntetos2005accuracy} proposed the \textit{Syntetos-Boylan Approximation} as an adjusted version of the \cite{croston1972forecasting} forecast formula.}
\end{eqfloat}

\begin{eqfloat}
\begin{equation}\label{EQ:TSB}
if d_t>0
\begin{cases}
    a_{t+1} = \alpha \cdot d_t + (1-\alpha)\cdot a_t\\
    p_{t+1} = \beta \cdot q \cdot d_t + (1-\beta)\cdot p_t\\
    f_{t+1} = a_{t+1} \cdot p_{t+1}
\end{cases}
if d_t=0
\begin{cases}
    a_{t+1} = a_t\\
    p_{t+1} = (1-\beta)\cdot p_t\\
    f_{t+1} = a_{t+1} \cdot p_{t+1}
\end{cases}
\end{equation}
\caption{Teunter, Syntetos \& Babai formula (\cite{teunter2011intermittent}) for irregular demand estimation, where \textit{a} is demand level, \textit{p} is probability of demand occurrence, \textit{d} refers to demand observations, \textit{q} is previous demand occurrence, and \textalpha represents a smoothing constant.}
\end{eqfloat}

A seminal work regarding intermittent demand forecasting was developed by \cite{croston1972forecasting}, who identified Exponential Smoothing as inadequate to estimate demand when the mean demand interval between two transactions greater than two time periods. Croston proposed a method to estimate the expected interval between transactions and the expected demand size (Eq.~\ref{EQ:CROSTON}), assuming that successive demand intervals and sizes are independent, the inter-demand intervals follow a Geometric distribution, and the demand sizes follow a Normal distribution. \cite{shenstone2005stochastic} showed Croston's method was not consistent with intermittent demand properties, but its results still outperformed conventional methods. Many researchers followed Croston's approach, either by enhancing this method or proposing similar ones. \cite{syntetos2005accuracy} proposed a slight modification to Croston's method, known as Syntetos-Boylan Approximation (SBA), to avoid a positive correlation between the forecasted demand size and the smoothing constant (Eq.~\ref{EQ:SBA}). \cite{leven2004inventory} suggested computing a new demand rate every time demand takes place, considering a maximum of one time per time bucket. \cite{teunter2011intermittent} consider computing a demand probability for each period and update the demand quantity forecast only when demand takes place (Eq.~\ref{EQ:TSB}). \cite{prestwich2014forecasting} proposed a hybrid of Croston's method and Bayesian inference to consider items obsolescence. \cite{chua2008short} developed an algorithm that estimates future demand occurrence based on three time-series: non-zero demands, the inter-arrival period between demand occurrences, and periods spanned between two demand occurrences. They estimate demand size with a simple moving average.

\subsubsection{Type II models}\label{RW-TYPE-II}
Type II models address irregular demand occurrence by performing data aggregation and achieving smooth time series. Much research was performed on the effect of aggregation on regular time series (\cite{hotta1993effect,souza2004effects,athanasopoulos2011tourism,rostami2013demand,petropoulos2015forecast,kourentzes2016forecasting}), showing that a higher aggregation improves forecast results.

\cite{nikolopoulos2011aggregate} proposed the aggregate-disaggregate intermittent demand approach (ADIDA), as a three-stage process: (i) perform time-series aggregation (either overlapping or non-overlapping aggregation), (ii) forecast the next time series value over the aggregated time series, and (iii) disaggregate the forecasted value to the original aggregation level.

\subsubsection{Type III models}\label{RW-TYPE-III}

\begin{figure*}[!ht]
\centering
\includegraphics[width=4.5in]{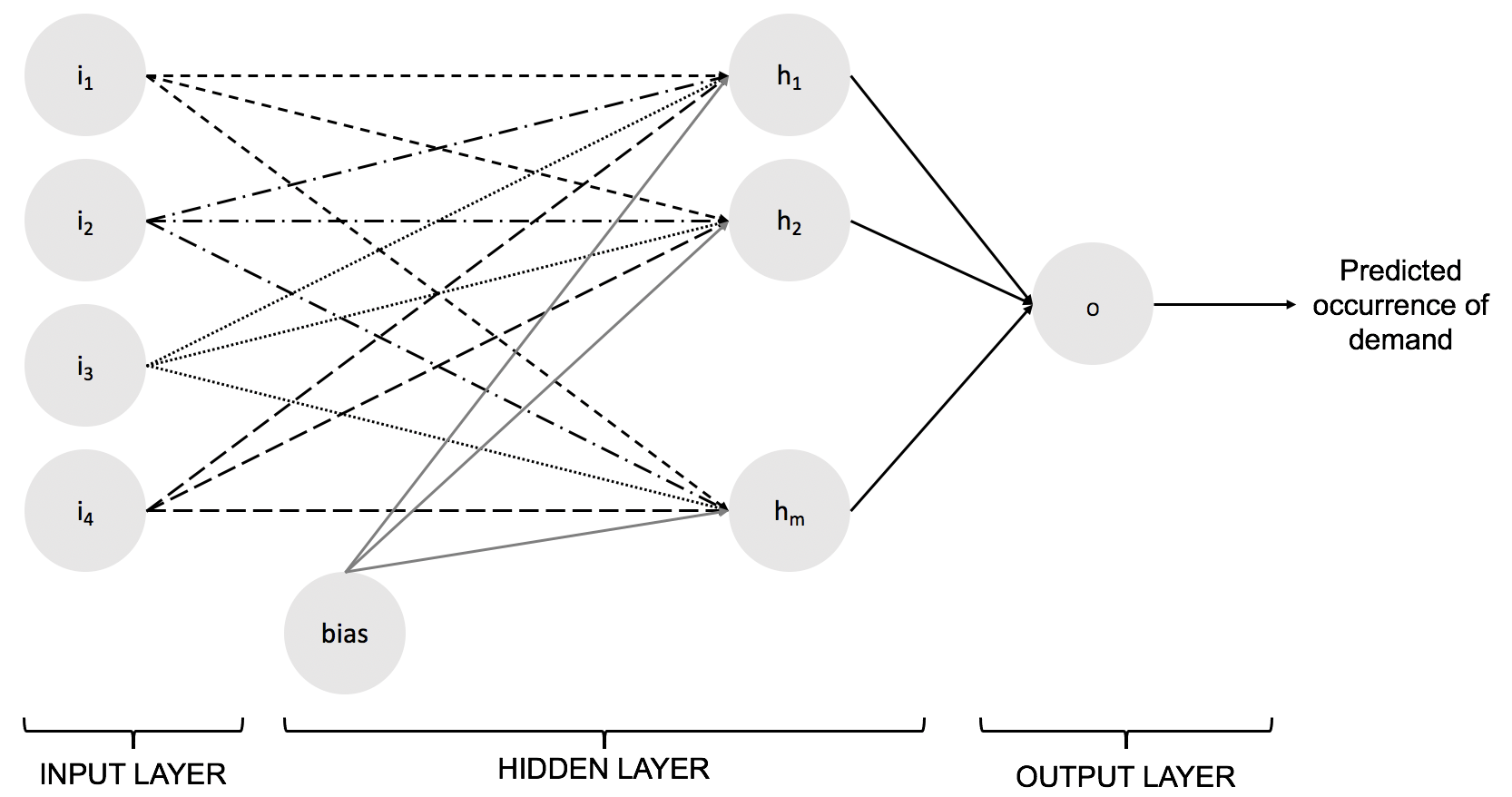}
\caption{MLP for hybrid approach proposed by \cite{nasiri2008hybrid}. Inputs to the model are demand size at the end of the preceding period, the number of periods between the last two demand occurrences, the number of periods between target period and last demand occurrence, and the number of periods between target period and first immediately preceding zero demand period.}
\label{F:HYBRID-MLP}
\end{figure*}

Following \cite{croston1972forecasting} intuition, some researchers developed separate models to forecast demand occurrence and demand size. \cite{willemain2004new} proposed to model demand occurrence as a Markov process and forecast demand size by randomly sampling past demand sizes and eventually jitter them to account for not yet seen values. \cite{hua2007new} follows a similar approach, attributing demand occurrence to autocorrelation or explanatory variables. If they attribute demand occurrence to autocorrelation, they predict demand occurrence based on Markov processes. Otherwise, they use a logistic regression model. \cite{nasiri2008hybrid} developed a hybrid approach, forecasting demand occurrence with a neural network (see Fig.~\ref{F:HYBRID-MLP}), while they estimate demand size with exponential smoothing. The neural network they propose considers four input variables: demand size at the end of the preceding period, the number of periods between the last two demand occurrences, number of periods between target period and last demand occurrence, and number of periods between target period and first immediately preceding zero demand period.
Finally, \cite{petropoulos2016another} developed an alternative perspective to the ADIDA framework (\cite{nikolopoulos2011aggregate}), aggregating time series in such a way that each time bucket contains a single demand occurrence. By doing so, the transformed time-series no longer present intermittency and forecast the time-varying number of periods when such demand will occur. Demand size is estimated based on Croston's, SBA, or Simple Exponential Smoothing methods, based on inter-demand interval and coefficient of variation mean values.

\subsubsection{Forecasting features}\label{RW-FORECASTING-FEATURES}
In the scientific literature related to demand forecasting of intermittent and lumpy demands, authors describe multiple characteristics and features relevant to demand occurrence forecasting. Among them we find the average inter-demand interval size (\cite{leven2004inventory}), previous demand event occurrence (\cite{gutierrez2008lumpy}), distribution of inter-demand interval sizes (\cite{croston1972forecasting}), demand size (\cite{nasiri2008hybrid}), demand shape distribution (\cite{zotteri2000impact}), usage of early information generated by customers during the purchasing process (\cite{verganti1997order}), the presence of paydays, billing cycles or holidays (\cite{hyndman2006another}), demand event autocorrelation (\cite{willemain1994forecasting}), or the fact that items may be purchased by same supplied or shipped using same transportation mode (\cite{syntetos2001forecasting}).

To estimate demand size, we found two techniques were applied across all cases. The first one was exponential smoothing (and its variants), applied across previous non-zero demand sizes (\cite{nasiri2008hybrid}). The second one was the use of jittering on top of randomly sampled past demand sizes to account for yet unseen values (\cite{willemain2004new,hua2007new}). \cite{altay2008adapting} described means to adjust demand size values based on the presence of trend in data.

\subsection{Metrics}\label{RW-METRICS}
Measurement of forecasting models performance for lumpy and intermittent demand was a subject of extensive research. \cite{syntetos2001forecasting} compared the performance of the Mean Signed Error, Wilcoxon Rank Sum Statistic, Mean Square Forecast Error, Relative Geometric Root Mean Square Error, and Percentage of times Better metrics. They concluded that the Relative Geometric Root Mean Square Error behaves well in the context of irregular demand. \cite{teunter2009forecasting} pointed out that the Relative Geometric Root Mean Square Error cannot be applied on a single item for zero or moving average forecasts (it would result in zero error). \cite{hemeimat2016forecasting} suggested using the tracking signal metric, calculated by dividing the most recent sum of forecast errors by the most recent estimate of Mean Absolute Deviation. Among many metrics, \cite{hyndman2006another} suggested using the MASE for lumpy and intermittent demands, providing a scale-free assessment of how accurate demand size forecasts are. Though traditional per-period forecasting metrics, such as Root Mean Squared Error, Mean Squared Error, Mean Absolute Deviation, or Mean Absolute Percentage Error, were widely used in the literature regarding irregular demand, \cite{teunter2009forecasting,kourentzes2014intermittent} showed they are not adequate due to the high proportion of zeros. \cite{prestwich2014mean} proposed computing a modified version of the error measures, considering the mean of the underlying stochastic process instead of the point demand for each point in time. Finally, it is relevant to point out that two metrics were used in the M5 competition (\cite{makridakis2020m5}) to assess time series regarding irregular sales: Root Mean Squared Scaled Error (RMSSE, introduced by \cite{hyndman2006another}), and the Weighted RMSSE. By using a score that considers squared errors, result measurements optimize towards the mean. The Weighted RMSSE variant allows for penalizing each time series error based on some criteria (e.g., item price). Both metrics, though, unevenly penalize products sold during the whole time-period against those that are not.

\cite{syntetos2010variance} noted that regardless of the metrics used to estimate how accurate demand forecasts are, it is crucial to measure the impact of forecasts on stock-holding and service levels. In this line, \cite{wallstrom2010evaluation} proposed two complementary metrics. The first one is the number of shortages, counting how many times the cumulated forecast error is over zero in the time interval of interest. The second one is Periods in Stock, as the number of periods the forecasted items spent in fictitious stock (or how many stock-out periods existed).
More recently, \cite{martin2020new} proposed the Stock-keeping-oriented Prediction Error Costs (SPEC), which considers for each time step demand forecasts translate either into costs of opportunity or stock-keeping costs, but never both at the same time.

\section{Reframing Demand Forecasting}\label{DEMAND-FORECASTING-REFRAMED}

\subsection{Demand characterization and forecasting models}\label{DFR-DEMAND-CHARACTERIZATION}

\begin{figure*}[!ht]
\centering
\includegraphics[width=5.0in]{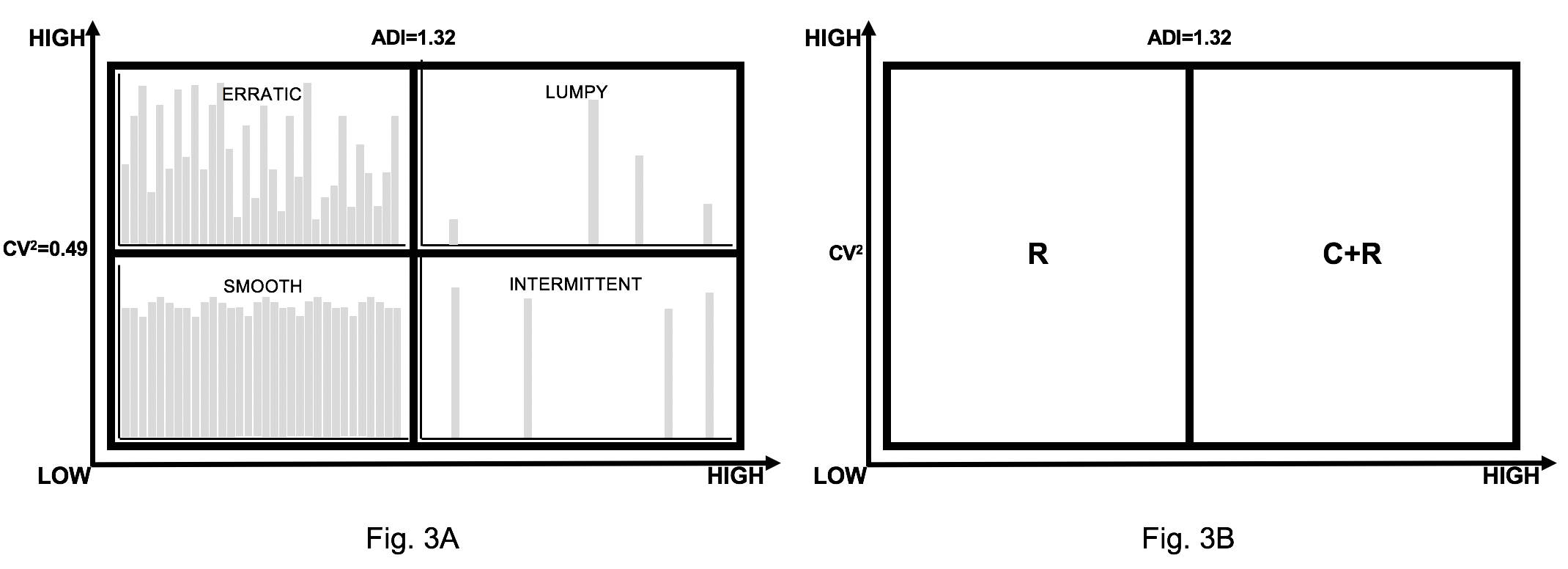}
\caption{Demand categorization schemas. On the right (Fig. 2A) corresponds to the influential categorization developed by \cite{syntetos2005categorization}. On the left (Fig. 2B), we propose a new schema that only considers demand occurrence, dividing demand into two groups. 'R' denotes regular demand occurrence, where demand size can be predicted with a regression model. 'C+R' denotes irregular demand occurrence that requires a model to predict demand occurrence and a model to predict demand size.}
\label{F:DEMAND-CLASSIFICATION-NEW}
\end{figure*}

\cite{croston1972forecasting} developed the intuition of considering two components for intermittent demand forecasts: demand occurrence and demand sizes. \cite{syntetos2005categorization} considered these two components and developed an influential demand categorization dividing demand into four types: smooth, erratic, intermittent, and lumpy, based on the coefficient of variation and the average demand interval.

Many authors followed Croston's intuition, developing separate models to estimate demand occurrence and demand size (e.g., \cite{willemain2004new,hua2007new,nasiri2008hybrid}), though none of them measured the performance of the demand occurrence component. We thus propose decoupling the demand forecasting problem into two sub-problems, each of which requires a separate model, features, and metrics: (i) demand occurrence, addressed as a classification problem, and (ii) demand size estimation addressed as a regression problem. Following the original work by \cite{syntetos2005categorization} and the division mentioned above, we propose an alternative demand categorization schema. Considering demand occurrence and demand quantity forecasting as two different problems, we can divide demand into two types (see Fig.~\ref{F:DEMAND-CLASSIFICATION-NEW}). The first demand type is 'R', which refers to demand with regular demand event occurrence. Since demand (almost) always occurs, the estimation of demand occurrence is rendered irrelevant and does not pose a challenge when developing a regression model to estimate demand size. The second demand type is 'C+R' and refers to demand with irregular demand event occurrence. These cases benefit from models that consider both demand occurrence and demand size (see Section~\ref{RW-DEMAND-FORECASTING-MODELS}).

\begin{figure*}[!ht]
\centering
\includegraphics[width=5.0in]{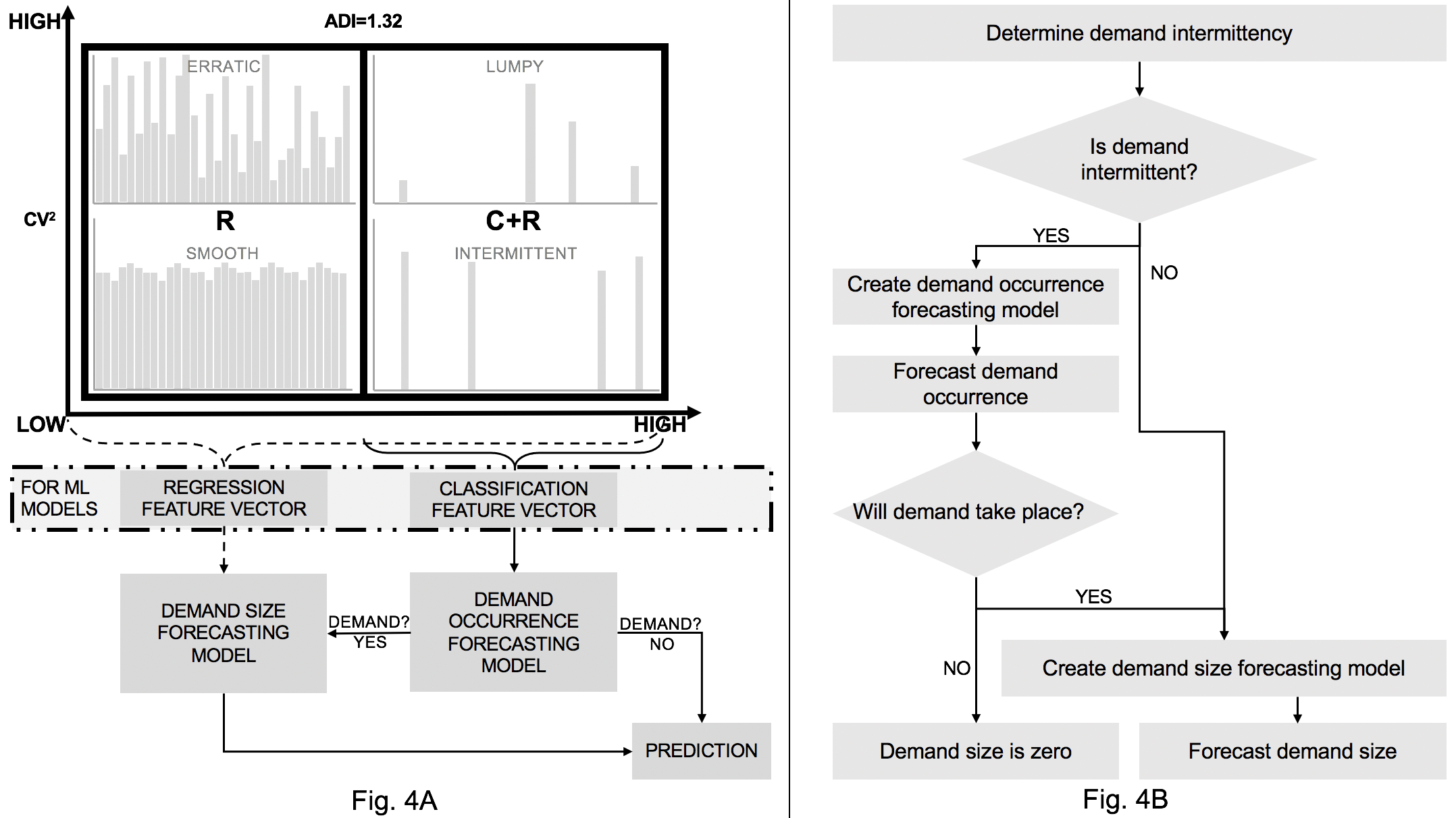}
\caption{Two-fold machine learning approach to demand forecasting. Fig. 4A shows a basic architecture for demand forecasting when reframing demand forecasting as classification and regression problems. Fig. 4B shows a fluxogram with steps followed to create the demand forecasting models and issue demand forecasts.}
\label{F:ML-MODEL-FLUXOGRAM}
\end{figure*}

We present a demand forecasting model architecture and a fluxogram describing how to build it and issue demand forecasts in Fig.~\ref{F:ML-MODEL-FLUXOGRAM}. Since the classification and regression models address different problems, we expect them to use different features to help achieve their goals. We described aspects relevant to demand occurrence and demand size forecasting present in the literature in Section~\ref{RW-FORECASTING-FEATURES}. These can be used as features for the classification and regression models, respectively.

An increasing body of research suggests global machine-learning time-series models (models built with multiple time series) provide better results than local ones (models considering time series corresponding to a single product) (\cite{bandara2020forecasting,salinas2020deepar})). Increased performance is observed even when training models with disparate time series have different magnitudes or may not be considered related to each other (\cite{laptev2017time}), though how to bound the maximum possible error in such models remains a topic of research. Given this insight, we consider that dividing demand based on the coefficient of variation provides limited value and is no longer relevant to demand categorization.

We keep the cut-off value of ADI=1.32 proposed by \cite{syntetos2005categorization} as a reference. While this cut-off value remains relevant for statistical methods, we consider its relevance to blur regarding different machine learning models. Its importance may be rendered irrelevant for global machine learning classification models that predict demand occurrence. By considering multiple items at once, global models perceive a higher density of demand events and less irregularity than models developed with data regarding a single demand item. Simultaneously, the model can learn underlying patterns, which may be related to specific behaviors (e.g., deliveries take place only on certain days). It is important to note that event scarcity usually results in imbalanced classification datasets, posing an additional challenge.

We present suggested metrics to assess each model, and overall demand forecasting performance, in Section~\ref{DFR-METRICS}.

\subsection{Metrics}\label{DFR-METRICS}

Though several authors (e.g., \cite{croston1972forecasting,syntetos2001bias,kieferdemand}) considered separating demand occurrence and demand size when forecasting irregular demand, in the literature we reviewed, we found no researcher to measured them separately. We thus consider they did not consider demand occurrence and demand size forecast as entirely different problems.

Considering irregular demand forecasting only as a regression problem lead to much research and discussion (presented in Section~\ref{RW-METRICS}) on how to mitigate and integrate zero-demand occurrence to measure demand forecasting models performance adequately. In our research, we provide a different perspective. We adopt four metrics to assess the performance of demand forecasting models: (i) AUC ROC to measure how accurate the model is forecasting demand occurrence, (ii) two variants of MASE to measure how accurate the model is forecasting demand size, and (iii) SPEC to measure how the forecast impacts inventory. When measuring SPEC, we consider \textalpha\textsubscript{1}=\textalpha\textsubscript{2}=0.5.


AUC ROC is widely adopted as a classification metric, having many desirable properties such as being threshold independent and invariant to a priori class probabilities. MASE has the desirable property of being scale-invariant. We consider two variants (namely MASE\textsubscript{I} and MASE\textsubscript{II}). Following the criteria in \cite{wallstrom2010evaluation}, we compute MASE\textsubscript{I} on the time series that results from ignoring zero-demand values. By doing so, we assess how well the regression model performs against a N\"aive forecast, assuming a perfect demand occurrence prediction.
On the other hand, MASE\textsubscript{II} is computed on time series considering all points where either demand took place or the classification model predicted demand occurrence. By doing so, we measure the impact of demand event occurrence misclassification on the demand size forecast. When the model predicting demand occurrence has perfect performance, (i) should equal (ii). Finally, we compute SPEC on the whole time series (considering zero and non-zero demand occurrences). This way, the metric measures the overall forecast impact on inventory, weighting stock-keeping, and opportunity costs.

\section{Methodology}\label{METHODOLOGY}
\subsection{Business understanding}\label{M-BUSINESS-UNDERSTANDING}
Demand forecasting is a critical component to supply chain management since its outcomes directly affect the supply chain and manufacturing plant organization. This research focuses on demand forecasting for a European original equipment manufacturer from the automotive industry. We explored providing demand forecasts for each material and client at a daily level. Such forecasts enable highly detailed planning.
From the forecasting perspective, accurate forecasts at a daily level can leverage the most recent information, which is lost at higher aggregation levels for non-overlapping aggregations. They also avoid imprecisions that result from higher-level forecast disaggregations.

\subsection{Data understanding}\label{M-DATA-UNDERSTANDING}

\begin{table*}[t]
\centering
\scalebox{0.80}{
\begin{tabular}{|r|r|r|r|r|r|r|r|}
\hline
metric & mean & std & min & 25\% & 50\% & 75\% & max \\ \hline
ADI & 86.00 & 87.26 & 1.97 & 11.86 & 37.29 & 156.6 & 261.00 \\ \hline
CV\textsuperscript{2} & 1.44 & 1.04 & 0.50 & 0.70 & 1.10 & 1.90 & 4.83 \\ \hline
\end{tabular}}
\caption{Summary statistics for 49 lumpy demand time series. \label{T:SUMMARY-STATS-LUMPY}}
\end{table*}

\begin{table*}[t]
\centering
\scalebox{0.85}{
\begin{tabular}{|r|r|r|r|r|r|r|r|}
\hline
metric & mean & std & min & 25\% & 50\% & 75\% & max \\ \hline
ADI & 56.72 & 70.58 & 1.41 & 9.79 & 25.26 & 71.18 & 261.00 \\ \hline
CV\textsuperscript{2} & 0.09 & 0.11 & 0.00 & 0.02 & 0.05 & 0.13 & 0.48 \\ \hline
\end{tabular}}
\caption{Summary statistics for 467 intermittent demand time series. \label{T:SUMMARY-STATS-INTERMITTENT}}
\end{table*}

For this research, we used a dataset with three years of demand data extracted from an Enterprise Resource Planning software. We consider records accounting products shipped from manufacturing locations for demand data on the day they are shipped.
The dataset comprises 516 time series that correspond to 279 materials and 149 clients. When categorized according to the schema proposed by \cite{syntetos2005categorization}, we found 49 correspond to a lumpy demand pattern, and 467 to intermittent demand pattern. In Table~Table~\ref{T:SUMMARY-STATS-LUMPY} and Table~\ref{T:SUMMARY-STATS-INTERMITTENT} we provide summary statistics for time series corresponding to each demand pattern. We find that demand occurrence for both sets of time series is highly infrequent, having a mean of one demand event in almost two months or more.

\subsection{Data Preparation, Feature Creation and Modeling}\label{M-DATA-PREPARATION}

\begin{figure*}[!ht]
\centering
\includegraphics[width=4.0in]{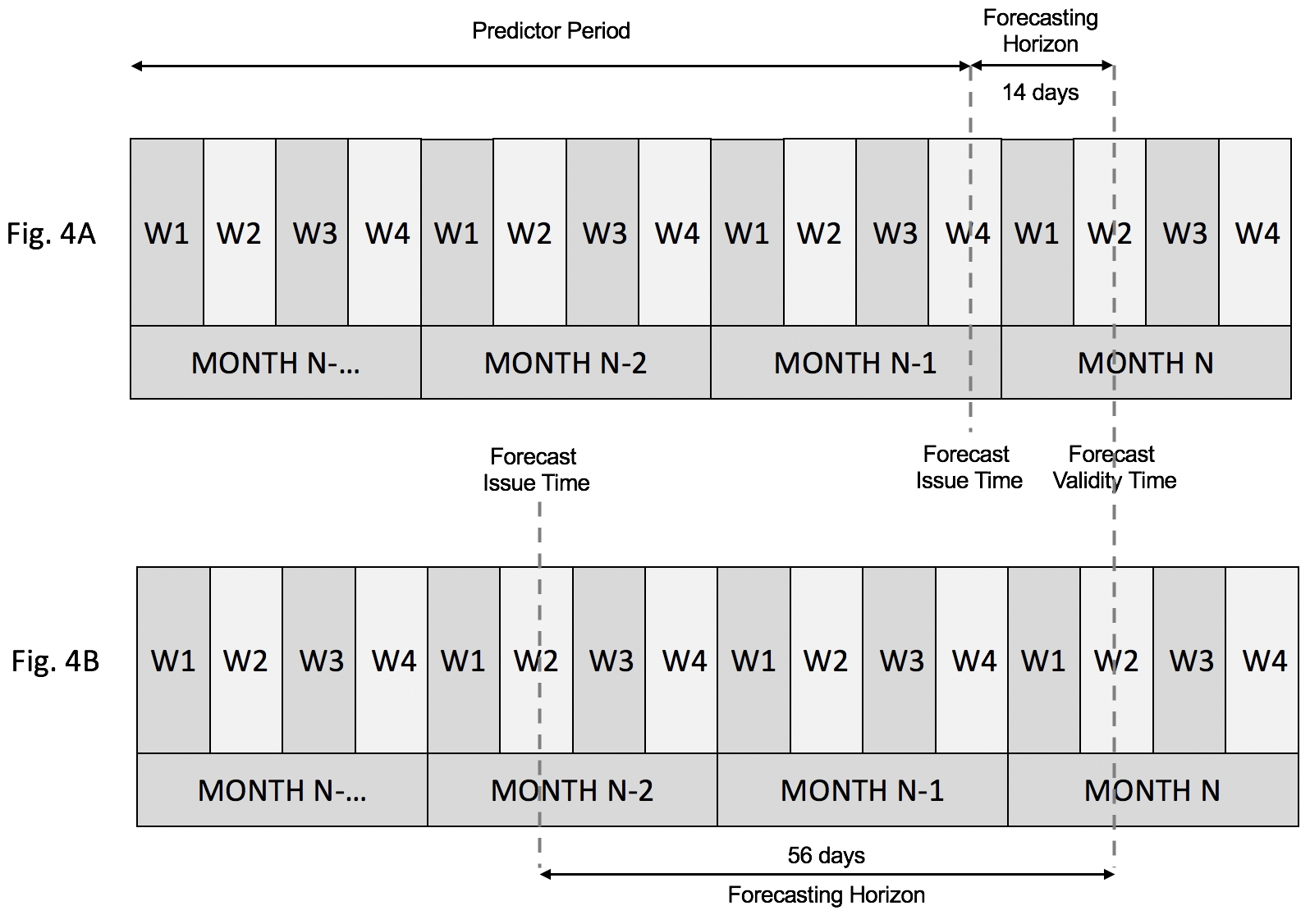}
\caption{We compute predictions for two forecasting horizons: 14 and 56 days, to test the sensitivity of predictions regarding demand occurrence and demand size to the forecasting horizon.}
\label{F:FORECASTING-PERIOD}
\end{figure*}

We forecast irregular demand with two separate models: a classification model to predict demand occurrence and a regression model to predict demand size. Though source data is the same for both, different features must address each model's goals. Following knowledge distilled in the literature, we created features presented in Section~\ref{RW-FORECASTING-FEATURES}, and some features of our own.

To model demand occurrence, we considered weekdays since last demand, day of the week of the last demand occurrence, day of the week for the target date, mean of inter-demand intervals, the mean of last inter-demand intervals (across all products), the skew and kurtosis of demand size distributions, among others.

To model demand size, we considered, for each product, the size of the last demand, the average of the last three demand occurrences, the median value of past occurrences, and the most frequent demand size value, the exponential smoothing of past values, among others.

When computing feature values, we considered two forecasting horizons: fourteen and fifty-six days (Fig.~\ref{F:FORECASTING-PERIOD}), to understand how the forecasting horizon size affects the forecasts.

To forecast irregular demand, we compare seven methods from the literature:
\begin{itemize}
    \item Croston's method (\cite{croston1972forecasting})
    \item SBA (\cite{syntetos2005accuracy})
    \item TSB (\cite{teunter2009bias})
    \item hybrid model proposed by \cite{nasiri2008hybrid}. Considers a NN model (see Fig.~\ref{F:HYBRID-MLP}) to forecast demand occurrence, while demand size is computed as exponential smoothing over non-zero demand quantities in past periods;
    \item a hybrid model proposed by \cite{willemain2004new}. Demand occurrence is estimated as a Markov process, while demand sizes are randomly sampled from previous occurrences;
    \item ADIDA forecasting method, proposed by \cite{nikolopoulos2011aggregate}, which removes intermittence through aggregation, and then dissagregates the forecast back to the original aggregation level.
\end{itemize}

We also developed models of our own. We created a Catboost model (\cite{prokhorenkova2018catboost}) to forecast demand occurrence and compare six models to forecast demand size: N\"aive, Most Frequent Value (MFV), Moving Average over last three demand periods (MA(3)), Simple Exponential Smoothing (SES), random sampling from past values with jittering (RAND), and Random Forest Regressor (ML). While we initially used the LightGBM algorithm (\cite{ke2017lightgbm}) since four of the top five time-series forecasting models in the M5 competition were based on this algorithm (\cite{makridakis2020m5}), we obtained better results with the Random Forest Regressor.

Catboost uses gradient descent to minimize a cost function, which informs how successful it is towards meeting the classification goal. Since the dataset regarding demand occurrence is heavily imbalanced (less than 6\% of instances correspond to demand occurrence), we choose to optimize the model training with the focal loss (\cite{lin2017focal}). The focal loss has the desirable property of providing an asymmetric penalization to train samples, focusing on misclassified ones to improve the overall classification.

\section{Experiments and Results}\label{EXPERIMENTS-AND-RESULTS}

This section describes the experiments we conducted and assesses their results with metrics we described in Section~\ref{DFR-METRICS}. For the SPEC metric, we considered \textalpha\textsubscript{1} and \textalpha\textsubscript{2} equal to 0.5. We summarize our experiments in Table~\ref{T:EXPERIMENTS-DESCRIPTION}, while we present their results in Table~\ref{T:EXPERIMENTS-RESULTS-ALL} and Table~\ref{T:EXPERIMENTS-RESULTS-COMPARISON}.

We adopt two forecasting horizons (fourteen and fifty-six days) to understand how sensitive are the existing approaches to the forecast lead time. To evaluate the models, we used nested cross-validation (\cite{stone1974cross}), which is frequently used to evaluate time-sensitive models. We test our models considering making predictions at the weekday level for six months of data. For classification models, we measure AUC ROC considering prediction scores cut at a threshold of 0.5. The only exception to this was the model by \cite{nasiri2008hybrid} since the author explicitly stated that they considered any prediction above zero as an indication of demand occurrence.

We observed that though the literature approaches provided different means to estimate demand occurrence, their performance to do so was close to 0.5 AUC ROC. Differences in performance were mainly driven by the method used to estimate demand size.

All the models we developed in our experiments strongly outperformed the models replicated from the literature. When considering AUC ROC, our models achieved scores of at least 0.94, almost doubling every model described in the literature. When considering regression metrics, our models displayed three to four times better MASE\textsubscript{I} and MASE\textsubscript{II}, and even more significant differences regarding the SPEC metric. We consider these results confirm the importance of considering the demand forecasting of irregular demands as two separate problems (demand occurrence and demand size), each with its features and optimized against its own set of metrics. Improvements in classification scores have a substantial impact on demand size and inventory metrics.

We obtained the best results with the Catboost classifier trained over all instances, making no difference between lumpy or intermittent demand. The model achieved an almost perfect classification AUC ROC score, reaching a value of 0.97 for fourteen and fifty-six days horizon. Among regression models used to estimate demand size, we achieved the best results with SES and MFV. The first one outperformed every other model on MASE\textsubscript{I} and MASE\textsubscript{II}, while the second one had the best median SPEC score and remained competitive on MASE\textsubscript{I} and MASE\textsubscript{II}.

When building the global classification model, we were interested in how much better it performed than models built using only lumpy and intermittent demands. We found that while the model built only on lumpy or intermittent demand achieved AUC ROC of at least 0,7368 and 0,9666 in each subset of products, while the global model built on all time series increased the performance to 0,9097 and 0,9776 respectively (see Table~\ref{T:EXPERIMENTS-ROC-GROUPS}).

From the results, it seems the forecasting horizon has little influence on the classification performance. We attribute this difference to the fact that demand occurrence is scarce, and thus changes in demand behavior are likely to be slow. This fact has engineering implications since there is no need to retrain and deploy the classification model frequently. Achieving the best demand size forecasts with SES and MFV also means the demand forecasting does not require expensive computations and much maintenance in production.

Finally, we compared our C2R1-SES model against the ADIDA approach. To that end, we selected 64 products that had steady demand over the three years we considered and used SES to provide two-months ahead forecasts and measured MASE for the last six months using cross-validation. The C2R1-SES model with a fifty-six day forecasting horizon showed a strong performance, achieving a MASE of 0,0052, while for ADIDA-SES, we measured a MASE of 1,6640. Our model outperforms the state-of-the-art self-improving mechanism ADIDA at an aggregate level. In the light of the results obtained, we consider our approach has at least two significant advantages over aggregate forecasts. First, we avoid issues related to forecasts disaggregation, provided the classification model is good enough. Second, our models benefit from the information that is lost with aggregation, which may help to issue better forecasts.

\begin{table*}[t]
\centering
\resizebox{\columnwidth}{!}{
\begin{tabular}{|l|l|l|l|l|}
\hline
\textbf{Model Task} & \textbf{Model Type} & \textbf{ID} & \textbf{Models} & \textbf{Data} \\ \hline
\multirow{2}{*}{\textbf{Classification}} & \begin{tabular}[c]{@{}l@{}}Global, per demand type\\ (lumpy, intermittent)\end{tabular} & C1 & Catboost & Demand occurrence features. \\ \cline{2-5}
 & \begin{tabular}[c]{@{}l@{}}Global, over all instances\\ (lumpy and intermittent)\end{tabular} & C2 & Catboost & Demand occurrence features. \\ \hline
\multirow{3}{*}{\textbf{Regression}} & Local, one per each time series. & R1 & \begin{tabular}[c]{@{}l@{}}Naive\\ SES\\ MA(3)\\ MFV\\ RAND\end{tabular} & Past non-zero demand sizes. \\ \cline{2-5}
 & \begin{tabular}[c]{@{}l@{}}Global, per demand type\\ (lumpy, intermittent)\end{tabular} & R2 & LightGBM & Demand size features. \\ \cline{2-5}
 & \begin{tabular}[c]{@{}l@{}}Global, over all instances \\ (lumpy and intermittent)\end{tabular} & R3 & LightGBM & Demand size features. \\ \hline
\end{tabular}
}
\caption{Description of reference models we evaluated for demand occurrence forecast, and demand size estimation. \label{T:EXPERIMENTS-DESCRIPTION}}
\end{table*}

\begin{table*}[t]
\centering
\resizebox{\columnwidth}{!}{
\begin{tabular}{|l|l|r|r|r|r|r|r|r|r|}
\hline
\multicolumn{1}{|c|}{\multirow{2}{*}{\textbf{Model}}} & \multicolumn{1}{c|}{\multirow{2}{*}{\textbf{Regression}}} & \multicolumn{4}{c|}{\textbf{Forecasting   Horizon: 14 days}}            & \multicolumn{4}{c|}{\textbf{Forecasting   Horizon: 56 days}}            \\ \cline{3-10}
\multicolumn{1}{|c|}{} & \multicolumn{1}{c|}{} & \textbf{AUC ROC} $\uparrow$ & \textbf{MASE\textsubscript{I}} $\downarrow$ & \textbf{MASE\textsubscript{II}} $\downarrow$ & \textbf{SPEC\textsubscript{median}} $\downarrow$   & \textbf{AUC ROC} $\uparrow$ & \textbf{MASE\textsubscript{I}} $\downarrow$ & \textbf{MASE\textsubscript{II}} $\downarrow$ & \textbf{SPEC\textsubscript{median}} $\downarrow$   \\ \hline
\textbf{C1R1}                                         & \textbf{Naive}                                            & 0,9408           & 0,5764          & 1,1084          & 121,9809         & \textbf{0,9409}  & 0,4861          & 1,1084          & 121,9809         \\ \hline
\textbf{C1R1}                                         & \textbf{MA(3)}                                            & 0,9408           & 0,5482          & 1,0544          & 281,2004         & \textbf{0,9409}  & 0,4624          & 1,0544          & 281,2004         \\ \hline
\textbf{C1R1}                                         & \textbf{SES}                                              & 0,9408           & 0,5229          & 1,0058          & 1210,7634        & \textbf{0,9409}  & 0,441           & 1,0058          & 1210,7634        \\ \hline
\textbf{C1R1}                                         & \textbf{MFV}                                              & 0,9408           & 0,5437          & 1,0457          & 141,4351         & \textbf{0,9409}  & 0,4585          & 1,0457          & 141,4351         \\ \hline
\textbf{C1R1}                                         & \textbf{RAND}                                             & 0,9408           & 0,7552          & 1,4524          & 2011,5744        & \textbf{0,9409}  & 0,6353          & 1,4485          & 2675,2844        \\ \hline
\textbf{C1R2}                                         & \textbf{ML}                                               & 0,9408           & 0,5813          & 1,1183          & 46422,3206       & \textbf{0,9409}  & 0,4917          & 1,1215          & 46162,9523       \\ \hline
\textbf{C1R3}                                         & \textbf{ML}                                               & 0,9408           & 0,5758          & 1,125           & 48807,4847       & \textbf{0,9409}  & 0,4613          & 1,1274          & 48279,0973       \\ \hline
\textbf{C2R1}                                         & \textbf{Naive}                                            & \textbf{0,9700}  & 0,5271          & 1,046           & 110,3435         & \textbf{0,9700}  & 0,4445          & 1,046           & 110,3435         \\ \hline
\textbf{C2R1}                                         & \textbf{MA(3)}                                            & \textbf{0,9700}  & 0,4906          & 0,9736          & 245,1851         & \textbf{0,9700}  & 0,4137          & 0,9736          & 245,1851         \\ \hline
\textbf{C2R1}                                         & \textbf{SES}                                              & \textbf{0,9700}  & \textbf{0,4611} & \textbf{0,9152} & 1343,2328        & \textbf{0,9700}  & \textbf{0,3888} & \textbf{0,9152} & 1343,2328        \\ \hline
\textbf{C2R1}                                         & \textbf{MFV}                                              & \textbf{0,9700}  & 0,476           & 0,9448          & \textbf{94,8092} & \textbf{0,9700}  & 0,4014          & 0,9448          & \textbf{94,8092} \\ \hline
\textbf{C2R1}                                         & \textbf{RAND}                                             & \textbf{0,9700}  & 0,7267          & 1,4422          & 2034,5172        & \textbf{0,9700}  & 0,5975          & 1,4059          & 2938,0534        \\ \hline
\textbf{C2R2}                                         & \textbf{ML}                                               & \textbf{0,9700}  & 0,5194          & 1,031           & 44255,2309       & \textbf{0,9700}  & 0,4398          & 1,0352          & 44101,6088       \\ \hline
\textbf{C2R3}                                         & \textbf{ML}                                               & \textbf{0,9700}  & 0,5295          & 1,051           & 39918,1584       & \textbf{0,9700}  & 0,4479          & 1,0542          & 40219,3378       \\ \hline
\end{tabular}
}
\caption{Overall results obtained with the models we proposed, for both forecasting horizons. \label{T:EXPERIMENTS-RESULTS-ALL}}
\end{table*}

\begin{table*}[t]
\centering
\resizebox{\columnwidth}{!}{
\begin{tabular}{|l|r|r|r|r|r|r|r|r|}
\hline
\multicolumn{1}{|c|}{\multirow{2}{*}{\textbf{Model}}} & \multicolumn{4}{c|}{\textbf{Forecasting   Horizon: 14 days}}            & \multicolumn{4}{c|}{\textbf{Forecasting   Horizon: 56 days}}            \\ \cline{2-9}
\multicolumn{1}{|c|}{}                                & \textbf{AUC ROC} $\uparrow$ & \textbf{MASE\textsubscript{I}} $\downarrow$  & \textbf{MASE\textsubscript{II}} $\downarrow$ & \textbf{MASE\textsubscript{median}} $\downarrow$   & \textbf{AUC ROC} $\uparrow$ & \textbf{MASE\textsubscript{I}} $\downarrow$  & \textbf{MASE\textsubscript{II}} $\downarrow$ & \textbf{SPEC\textsubscript{median}} $\downarrow$   \\ \hline
\textbf{\cite{croston1972forecasting}}                                               & 0,5000           & 1,5997          & 96,2732         & 182590225,2842   & 0,5000           & 1,4769          & 1,4769          & 173846999,3033   \\ \hline
\textbf{\cite{syntetos2005accuracy}}                                                   & 0,5000           & 1,5196          & 91,4593         & 173457612,1803   & 0,5000           & 1,4047          & 88,2762         & 165151599,2596   \\ \hline
\textbf{\cite{teunter2009bias}}                                                   & 0,5337           & 1,0340          & 42,8525         & 82848841,2377    & 0,5448           & 0,7491          & 11,9425         & 20024649,5191    \\ \hline
\textbf{\cite{willemain2004new}}                                           & 0,5000           & 0,8616          & 1,3786          & 199,3702         & 0,5000           & 0,8619          & 1,3786          & 199,3702         \\ \hline
\textbf{\cite{nasiri2008hybrid}}                                           & 0,5000              & 0,5897          & 24,8423         & 1714,9372        & 0,5000              & 0,5906          & 24,8197         & 1714,9372        \\ \hline
\textbf{C2R1-SES}                                     & \textbf{0,9700}  & \textbf{0,4611} & \textbf{0,9152} & 1343,2328        & \textbf{0,9700}  & \textbf{0,3888} & \textbf{0,9152} & 1343,2328        \\ \hline
\textbf{C2R1-MFV}                                     & \textbf{0,9700}  & 0,4760           & 0,9448          & \textbf{94,8092} & \textbf{0,9700}  & 0,4014          & 0,9448          & \textbf{94,8092} \\ \hline
\end{tabular}}
\caption{Comparison of methods found in related work, and two of best models we created: C2R1-SES displays best performance on AUC ROC, MASE\textsubscript{I} and MASE\textsubscript{II}, while C2R1-MFV has best performance on the SPEC metric, while remaining competitive on the rest. \label{T:EXPERIMENTS-RESULTS-COMPARISON}}
\end{table*}

\begin{table*}[t]
\centering
\resizebox{\columnwidth}{!}{
\begin{tabular}{|l|r|r|r|r|}
\hline
\multicolumn{1}{|c|}{\multirow{2}{*}{\textbf{Model}}} & \multicolumn{2}{c|}{\textbf{Forecasting   Horizon: 14 days}} & \multicolumn{2}{c|}{\textbf{Forecasting   Horizon: 56 days}} \\ \cline{2-5}
\multicolumn{1}{|c|}{}                                & \textbf{AUC ROC\textsubscript{lumpy}} $\uparrow$   & \textbf{AUC ROC\textsubscript{intermittent}} $\uparrow$   & \textbf{AUC ROC\textsubscript{lumpy}} $\uparrow$    & \textbf{AUC ROC\textsubscript{intermittent}} $\uparrow$   \\ \hline
\textbf{C1}                                           & 0,7368                    & 0,9666                           & 0,7379                    & 0,9666                           \\ \hline
\textbf{C2}                                           & 0,9097                    & 0,9776                           & 0,9097                    & 0,9776                           \\ \hline
\end{tabular}}
\caption{Comparison of AUC ROC for lumpy and intermittent demand, obtained from C1 and C2 models. Using all data for a single classification model to predict demand occurrence shows improvements in both groups and time horizons. The highest improvement is observed for lumpy demand, with an improvement greater to 0,17.\label{T:EXPERIMENTS-ROC-GROUPS}}
\end{table*}

\section{Conclusions}\label{CONCLUSIONS}
In this research, we propose a new look at the demand forecasting problem for infrequent demands. Breaking it down into two prediction problems (classification for demand occurrence and regression for demand size), we (i) enable accurate model diagnostics and (ii) optimize each model for the specific task. Our results show that such decomposition enhances overall forecasting performance. Analyzing models proposed in the literature, we found that most of them underperformed by first failing to predict demand occurrence accurately. Together with the problem decomposition, we propose a set of four metrics (AUC ROC for classification, MASE\textsubscript{I} and MASE\textsubscript{II} for regression, and SPEC to assess the impact on inventory). We also developed a novel model, which outperformed six models described in the literature for lumpy and intermittent demand and the state-of-the-art self-improving mechanism known as ADIDA. Considering the problem separation mentioned above, we also propose a new demand classification schema based on the approach that provides a good demand forecast. We consider two types of demands: 'R' for demands where only regression is required and 'C+R' where classification and regression models estimate demand occurrence and demand size, respectively.

We envision several directions of future research. First, we would like to replicate these experiments on some widely cited datasets, such as SKUs of the automotive industry (initially used by \cite{syntetos2005accuracy}) or SKUs of Royal Air Force (used initially by \cite{teunter2009forecasting}). Second, we enable new approaches to deal with forecasts involving items obsolescence, stationarity, and trend on irregular demands by decoupling demand occurrence from demand size. Third, we understand this approach can be extended to forecasting irregular and infrequent sales, which can significantly impact retail. Finally, we consider this approach can be applied in other domains with irregular time series, where currently self-improving mechanisms, such as ADIDA, are applied.

\section*{Acknowledgement}
This work was supported by the Slovenian Research Agency and European Union’s Horizon 2020 program project FACTLOG under grant agreement number H2020-869951.





\bibliographystyle{elsarticle-harv}
\bibliography{main}

\end{document}